\documentclass[letterpaper, 10 pt, conference]{ieeeconf}  % Comment this line out if you need a4paper

\IEEEoverridecommandlockouts                              % This command is only needed if 
\usepackage{graphics} % for pdf, bitmapped graphics files
\usepackage{epsfig} % for postscript graphics files
\usepackage{mathptmx} % assumes new font selection scheme installed
\usepackage{times} % assumes new font selection scheme installed
\usepackage{amsmath} % assumes amsmath package installed
\usepackage{amssymb}  % assumes amsmath package installed
\usepackage{subfigure}
\usepackage{todonotes}
\usepackage{multirow}

\usepackage{subfloat}

\usepackage{breqn}
\usepackage{graphicx}

\usepackage{hyperref}
\hypersetup{colorlinks=true,urlcolor=blue,citecolor=red}

\usepackage{caption}
\newcommand{\etal}{\textit{et al}. }

\mathchardef\mhyphen="2D % Define a "math hyphen"

\title{\LARGE \bf
EV-IMO: Motion Segmentation Dataset \\and Learning Pipeline for Event Cameras
}

\author{Anton Mitrokhin$^{1*}$, Chengxi Ye$^{1*}$, \\
Cornelia Ferm\"uller$^{1}$, Yiannis Aloimonos$^{1}$, Tobi Delbruck$^{2}$%, Brianna Cash$^{3}$% <-this % stops a space
\thanks{$^{*}$The authors contribute equally to this work.}
\thanks{$^{1}$University of Maryland Institute for
Advanced Computer Studies,         
College Park, MD 20742, USA.
        {\tt\small E-mails: amitrokh@umd.edu, cxy@umd.edu, fer@umiacs.umd.edu, yiannis@cs.umd.edu}}%
\thanks{$^{2}$Department of Neuroinformatics, ETH Zurich, Z\"urich 8092, Switzerland.
        {\tt\small E-mail: tobi@ini.phys.ethz.ch}}%
}

\begin{document}

\maketitle
\thispagestyle{empty}
\pagestyle{empty}

%%%%%%%%% ABSTRACT
\begin{abstract}
%The last major challenge in research on visual navigation  is to accurately detect moving objects and estimate their motion.
%We present the first event-based learning approach and dataset (called EV-IMO) for moving object estimation and segmentation in indoor scenes.
% The problem is formulated in the SfM learning framework using a low parameter NN architecture on event data and extended to obtain in addition to camera pose  and depth estimation, also moving objects and   their  3D motion estimates. 
   %Our network consists of components, which estimate in  supervised mode the depth and probability for independent movement and in unsupervised mode the camera pose, which feed into a  component that  estimates residual object 3D motion. 
  %The approach uses both unsupervised and supervised components in a multi-level  architecture, and takes advantage of the idiosyncratic structure of the data. 
   %The work also introduces the first event-based dataset for independent motion detection -- featuring  real and  synthetic recordings of indoor sequences, with accurate ground truth depth and motion and object masks. Experiments show that the method has  high accuracy in depth estimation and camera segmentation, making it well suited for scene constrained robotics applications, and it has good performance on camera and object pose estimation.
%We present the first event-based learning approach and dataset -- \textit{EV-IMO} -- for segmentation of independently  moving  objects and estimation of their motion in indoor  scenes.

We present the first event-based learning approach for motion segmentation in indoor scenes and the first event-based dataset -- \textit{EV-IMO} -- which includes accurate pixel-wise motion masks, egomotion and ground truth depth.
Our approach is based on an efficient implementation of the SfM learning pipeline using a low parameter neural network architecture on event data. In addition to  camera egomotion and a dense depth map, the network estimates independently moving object segmentation  at the pixel-level and  computes per-object 3D translational velocities of moving objects. We also train a shallow network with just 40k parameters, which is able to compute depth and egomotion.

%In addition to camera pose and a dense depth map, the network estimates pose masks and pose vectors for the moving objects in a mixture model formulation.

%We introduce an efficient implementation of the SfM pipeline using a low-parameter  architecture and working on event stream of the DVS camera only. 
%We extend this network to predict candidate pose vectors, and use pose masks to assign these poses to each moving object. 
%As a result, our pipeline is capable of producing a pixel-wise independently moving object segmentation and compute per-object 3D translational velocities. 
%As a by-product, our pipeline outputs dense depth maps and 6 dof camera egomotion. 
%Furthermore, we train a shallow network with just 40k parameters, which is able to compute depth and egomotion.

Our EV-IMO dataset features 32 minutes of indoor recording with up to 3 fast moving objects in the camera field of view. 
The objects and the camera are tracked using a  VICON\textsuperscript{\textregistered} motion capture system. By 3D scanning the room and the objects, ground truth of the  depth  map  and  pixel-wise object masks are obtained. We then train and evaluate our learning pipeline on EV-IMO and demonstrate that it is well suited for scene constrained robotics applications.

%We present the first event-based learning approach and   dataset   –EV-IMO for   pixel-wise   segmentation   of   the independently  moving  objects  in  indoor  scenes. We introduce an efficient implementation of the SfM pipeline using a low-parameter  architecture and working on event stream of the DVS camera only. We extend this network to predict candidate pose vectors, and use pose masks to assign these poses to each moving object. As a result, our pipeline is capable of producing a pixel-wise independently moving object segmentation and compute per-object 3D translational velocities. As a by-product, our pipeline outputs dense depth maps and 6 dof camera egomotion. Additionally, we train a shallow network with just 40k parameters, which is able to compute depth and egomotion.

%Our EV-IMO dataset features 32 minutes of indoor recording with from 1 to 3 moving objects simultaneously on the camera frame. The objects and the camera are tracked by the VICON\textsuperscript{\textregistered} motion capture system. We use 3D scans of the room and objects to compute accurate depth map ground truth and pixel-wise object masks, which is reliable even in poor lighting conditions and during fast motion. We then evaluate our learning pipeline on EV-IMO and demonstrate that our approach far surpasses its rivals and is well suited for scene constrained robotics applications.

\end{abstract}

\section*{Supplementary Material}

The supplementary video, code, trained models and appendix will be made available at \url{http://prg.cs.umd.edu/EV-IMO.html}.

The dataset is available at \url{https://better-flow.github.io/evimo/}.

%%%%%%%%% BODY TEXT
\section{Introduction}
In modern mobile robotics, autonomous agents are often found in unconstrained, highly dynamic environments, having to quickly navigate around humans or other moving robots. This renders the classical structure from motion (SfM) pipeline, often implemented via SLAM-like algorithms, not only inefficient but also incapable of solving the problem of navigation and obstacle avoidance. An autonomous mobile robot should be able to instantly detect every independently moving object in the scene, estimate the distance to it and predict its trajectory, while at the same time being aware of its own egomotion.

In this light, event-based processing has long been of interest to computational neuroscientists, and a new type of imaging device, known as ``a silicon retina", has been developed by the neuromorphic community. The event-based sensor does not record image frames, but asynchronous temporal changes in the scene in form of a continuous stream of events, each of which is generated when a given pixel detects a change in log light intensity. This allows the sensor to literally \textit{see the motion} in the scene and makes it indispensable for motion processing and segmentation. The unique properties of event-based sensors - high dynamic range, high temporal resolution, low latency and high bandwidth allow these devices to function in the most challenging lighting conditions (such as almost complete darkness), while consuming a small amount of power.

\begin{figure}[t!]
\begin{center}
\includegraphics[width=1.0\columnwidth]{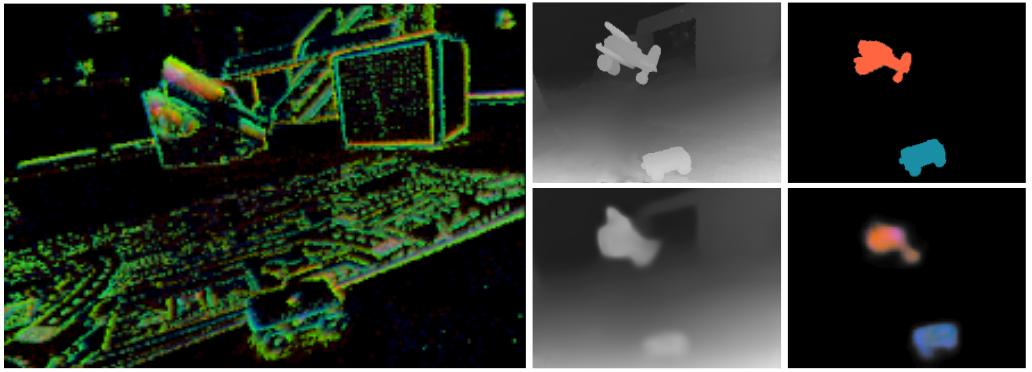}
\end{center}
\vspace{-0.5\baselineskip}
   \caption[Depth and per-pixel pose inference on sparse event data, on our \textit{EV-IMO} dataset]{\textit{\small{Depth and per-pixel pose inference on sparse event data, on our \textit{EV-IMO} dataset. The top row is the ground truth depth and pose (the color corresponds to the objects' linear velocities), the bottom row is the predicted network output. Camera egomotion is also estimated but not visualized here. Best viewed in color.}}}
   \vspace{-1.0\baselineskip}
\label{fig:teaser}
\end{figure}

We believe that independent motion detection and estimation is an ideal application for event-based sensors, especially when applied to problems of autonomous robotics. Compared to  classical cameras, event-based sensors encode spatio-temporal motion of image contours by producing a sparse data stream, which allows them to perceive extremely fast motion without experiencing motion blur. This, together with high tolerance to poor lighting conditions make this sensor a perfect fit for agile robots (such as quadrotors) which require a robust low latency visual pipeline.

On the algorithmic side, the estimation of 3D motion and scene geometry has been of great interest in Computer Vision and Robotics for quite a long time, but  most works considered the  scene to be static. Earlier classical algorithms studied the problem of Structure from Motion (SfM) \cite{fermuller2000observability} to develop  ``scene independent" constraints (e.g. the epipolar constraint \cite{hartley2003multiple} or depth positivity constraint \cite{fermuller1995passive})
and estimate 3D motion from images to facilitate subsequent scene reconstruction \cite{fermuller1997visual}. In recent years, most works have adopted the SLAM philosophy \cite{davison2007monoslam}, where depth, 3D motion and image measurements are estimated together using iterative probabilistic algorithms. Such reconstruction approaches are known to be computationally heavy and often fail in the presence of outliers.

To move away from the restrictions imposed by the classical visual geometry approaches, the Computer Vision and Robotics community started to lean towards learning. Yet, while the problem of detecting moving objects has been studied both in the model-based and learning-based formulation, estimating  object motion in addition to spatio-temporal scene reconstruction is still largely unexplored. An exception is the work in \cite{vijayanarasimhan2017sfmnet}, which however does not provide an evaluation. %In this paper we propose a learning-based solution to this problem using as input event-based data.

%Each event, $e(x, y, t, p)$ is encoded  by its pixel position  $x, y$, timestamp, $t$, accurate to microseconds, and polarity, $p \in \{−1, 1\}$, indicating whether the intensity decreased or increased. 

%In comparison to video, the data from this sensor, the event clouds,  are  much sparser  than sequences of images. Conceptually, event clouds encode the cues of motion and contour. This is a good fit for the problem of motion segmentation, which requires ``image motion'' information to encode spatio-temporal geometry, and some ``boundary'' information for grouping  the moving objects. Thus, it appears that DVS provides the minimal information necessary for motion segmentation, but with low latency and high dynamic range. DVS, thus, could be a real asset in solving the problem of motion segmentation under even challenging conditions.

%\todo{Motivational example}
%<Here we put the motivating example.>
%<IROS 2018 paper. Motion compensation alone does not allow for accurate segmentation>

%such as DAVIS \cite{DAVIS}.

In this work we introduce a compositional neural network (NN) pipeline, which provides supervised up-to-scale depth and pixel-wise motion segmentation of the scene, as well as \textit{unsupervised} 6 dof egomotion estimation and a per-segment linear velocity estimation using only monocular event data (see Fig. \ref{fig:teaser}). This pipeline can be used in indoor scenarios for motion estimation and obstacle avoidance.

We also created a dataset, \textit{EV-IMO}, which includes 32 minutes of indoor recording with multiple independently moving objects shot against a varying set of backgrounds and featuring different camera and object motions. To our knowledge, this is the first dataset for event-based cameras to include accurate pixel-wise masks of independently moving objects, apart from depth and trajectory ground truths.

To summarize, the contributions of this work are:

\begin{itemize}
    \item The first NN for estimating both camera and object 3D motion using event data;
    \item The first dataset -- \textit{EV-IMO} -- for motion segmentation with ground truth depth, per-object mask, camera and object motion;
%    \item The first NN architecture taking advantage of the idiosyncratic nature of event clouds ???
    \item A novel loss function tailored for event alignment, measuring the profile sharpness of the motion compensated events;
    \item Demonstration of the feasibility of using a shallow low parameter multi-level feature NN architecture for event-based segmentation while retaining similar performance with the full-sized network.
\end{itemize}

\section{Related Work}
\subsection{Event based Optical Flow, Depth and Motion Estimation}
Many of the first  event-based algorithms were concerned with  optical flow. Different techniques  employed the concepts of gradient computation \cite{benosman_asynchronous_2012,barranco2014contour,benosman_event_2014,mueggler2015lifetime}, template matching   \cite{liu2017block}, and frequency computation  \cite{barranco2015bio,brosch2015event} on event surfaces.
%Benosman \etal~\cite{benosman_asynchronous_2012} developed a local gradient based method, while Barranco \etal~\cite{barranco2014contour} proposed to obtain flow from spatial and temporal changes by rudimentary tracking  events over multiple time intervals. In subsequent research Benosman \etal and Mueggler \etal~\cite{benosman_event_2014,mueggler2015lifetime} fit local planes to time stamps in local spatial neighborhoods, and Orchard \etal~\cite{orchard2013spiking} provides a spiking neural network implementation based on that approach. Some works demonstrate that matching approaches e.g. \cite{liu2017block}, and frequency based techniques \cite{barranco2015bio,brosch2015event} can be used in event-based motion estimation. 
Zhu \etal~\cite{zhu2018ev} proposed a self-supervised deep learning approach using the intensity signal from the DAVIS camera for supervision.
%, and \cite{ye2018unsupervised} proposed a completely self-supervised network for egomotion and depth and optical flow estimation in self driving scenarios.

The problem of 3D motion  estimation was studied following the  visual odometry and SLAM formulation for the case of  rotation only \cite{reinbacher2017real}, with known maps \cite{weikersdorfer2013simultaneous,censi2013low,GallegoLMRDS16}, by combining event-based data with image measurements \cite{kueng2016low,tedaldi2016feature}, and using IMU sensors \cite{zhu2017event}. Other recent approaches jointly reconstruct the image intensity of the scene, and estimate 3D motion. First, in \cite{kim2014simultaneous} only rotation was considered, and in  \cite{kim2016real}  the   general case was addressed. 
%\subsection{Independent Motion Detection}
\subsection{Independent Motion Detection}
Many motion segmentation methods used in video applications are based on 2D measurements only \cite{sun2012layered,odobez1995mrf}. 3D approaches, such as the one here,  model the camera's rigid motion. Thompson  and Pong \cite{Thompson1990} first suggested detecting moving objects by checking contradictions to the epipolar constraint.  Vidal \etal \cite{vidal2003generalized} introduced the concept of subspace constraints for segmenting multiple objects. A good motion segmentation requires both constraints imposed by the camera motion and some form of scene constraints for clustering into regions. The latter can be achieved using approximate models of the rigid flow or the scene in view, for example by modeling the scene as planar, fitting multiple planes using the plane plus parallax constraint \cite{irani1998unified}, or selecting models depending on the scene complexity \cite{torr1998geometric}. In addition constraints on the occlusion regions \cite{ogale2005motion} and discontinuities \cite{fragkiadaki2012video}  have been used. Recently, machine learning techniques have been used for motion segmentation \cite{Fragkiadaki_2015_CVPR, bideau2018best}. As  discussed next, the well-known SfM learner acquires both, the depth map and the rigid camera motion,
and thus the flow due to  rigid motion is fully constrained.
\subsection{Learning in Structure from Motion}
In pioneering work, Saxena \etal~\cite{saxena2006learning} demonstrated that shape can be learned from single images, inspiring  many other supervised depth learning approaches (e.g.~\cite{eigen2014depth}). The concept was recently  adopted  in the SfM pipeline, and used  in stereo \cite{garg2016unsupervised} and video \cite{xie2016deep3d}.
%Tateno \etal~\cite{tateno2017cnn} predict the depth of key frames.  Garg \etal~\cite{garg2016unsupervised} and Xie \etal~\cite{xie2016deep3d} introduced self-supervised learning of depth using stereo and video, respectively.
Most recently, Zhou \etal~\cite{zhou2017unsupervised} took it a step further, and showed how to estimate 3D motion and depth through the supervision of optical flow. Wang \etal~\cite{wang2018learning},  instead of predicting depth in a separate network component, propose to incorporate a Direct Visual Odometry (DVO) pose predictor.  Mahjourian \etal~\cite{mahjourian2018unsupervised}  in addition to  image alignment  enforce  alignment of the geometric scene structure in the loss function. Yang \etal~\cite{Yang-Lego} added a 3D smoothness prior to the pipeline, which enables joint estimation of edges and 3D scene. Yin \etal~\cite{geonet} include a non-rigid motion localization component to also detect moving objects. Our architecture is most  closely  related to SfM-Net \cite{vijayanarasimhan2017sfmnet}, which learns using supervised and non-supervised components, depth, 3D camera and object motion. However, due to the lack of a dataset, the authors did not evaluate the object motion estimation.
Finally, there are two related studies in the literature on  event-based data: Zhu \etal \cite{zhu2017event} proposed the first unsupervised learning approach and applied it for optical flow estimation using the DAVIS sensor, where the supervision signal comes from the image component of the sensor. The arXiv paper \cite{ye2018unsupervised} first adopted the full structure from motion pipeline.
%of \cite{zhou2017unsupervised}.
Different from our work, this paper, like \cite{zhu2017event}, does not take advantage of the structure of event clouds. Most important, our work also detects, segments, and estimates the 3D motion of independently moving objects, and provides the means for evaluation.

\begin{figure*}
\begin{center}
    \includegraphics[width=0.99\textwidth]{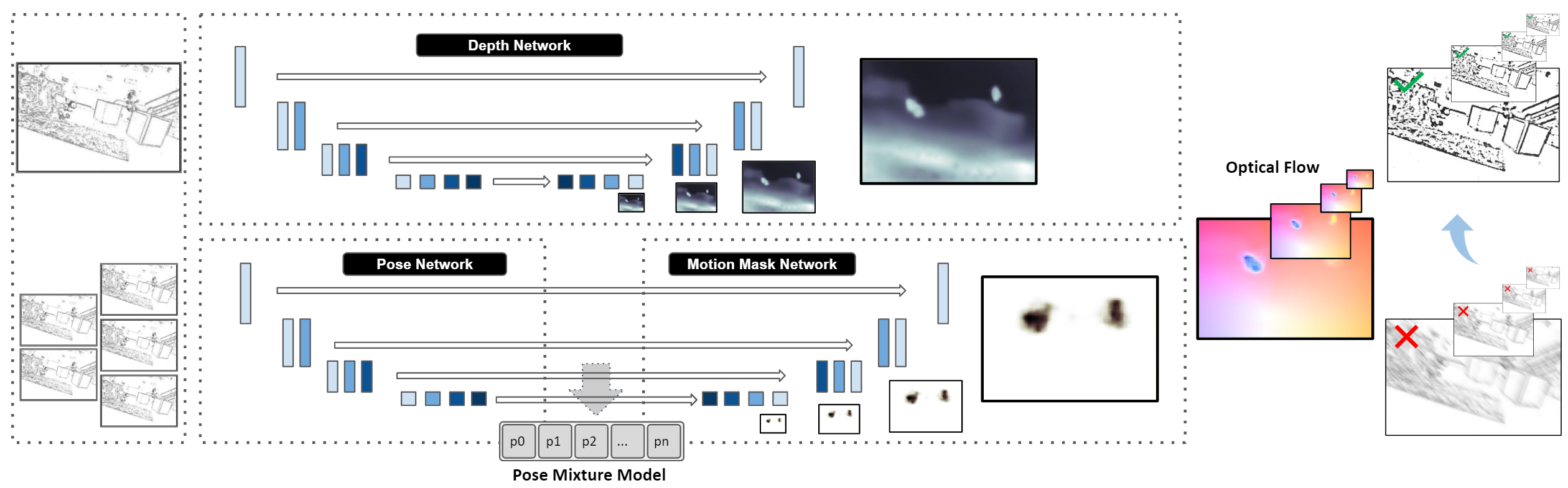}
\end{center}
\vspace{-1.0\baselineskip}
\caption[Our depth and motion estimation pipeline]{\textit{\small{A depth network (top) uses an encoder-decoder architecture and is trained in supervised mode to estimate scene depth. A pose network (bottom left) takes consecutive event slices to generate a mixture model for the pixel-wise pose. A mixture of poses and mixture probabilities (bottom right) are outputs of this network. The outputs of the two networks are combined to generate the optical flow, then to inversely warp the inputs and backpropagate the error. Our networks utilize multi-level feature representations (shown in different darkness) for the underlying tasks.}}}
\label{fig:pipeline}
\end{figure*}

\section{The Architecture}
\subsection{Network Input}
The raw data from the Dynamic Vision Sensor (DVS) is a continuous stream of events. Each event, $e(x, y, t, p)$ is encoded by its pixel position  $(x, y)$, timestamp $t$, accurate to microseconds, and binary polarity, $p \in \{ -1, 1\}$, indicating whether the intensity of light decreased or increased.

In  $(x, y, t)$ space, the event stream represents a 3D pointcloud. To leverage  maximum information from this representation and pass it down to the network, we subdivide the event stream into consecutive \textit{time slices} of size $\delta t$ (in our implementation - 25 $ms$). Every time slice is projected on a plane with a representation similar to our previous work \cite{ye2018unsupervised} - we create a 3 channel map, with 2 channels being positive and negative event counts, and one channel being a timestamp aggregate, as first proposed in \cite{iROSBetterFlow}. 

We then feed these 2D maps to the neural networks  in our pipeline. The benefit of the 2D input representation is the reduction of data sparsity, and a resulting increase in efficiency compared to the 3D learning approaches. Yet, the 2D input may suffer from motion blur during fast motions. We tackle this problem by using a fine scale warping loss (sec. \ref{sec:wloss}), which uses 1 $ms.$ slices to compute the loss.

\subsection{Overview  of the Architecture}

 Our pipeline (see Fig. \ref{fig:pipeline}) consists of a depth prediction network and a
 pose prediction network. Both networks are low parameter~\cite{ye2018unsupervised} encoder-decoder networks~\cite{DBLP:conf/miccai/RonnebergerFB15}. Our depth network  performs prediction on a single slice map. A supervision loss $Loss_{depth}$ comes by comparing with the ground truth as we describe in subsection~\ref{sec:DepthLoss}. Our pose network uses up to 5 consecutive maps, to better account for the 3D structure of the raw event data. The pose network utilizes a mixture model to estimate pixel-wise 3D motion (relative pose) and corresponding motion masks from consecutive event slices. The masks are learned in supervised mode. We introduce a $Loss_{mask}$ on the motion mask. Finally, the two network outputs are used to generate the optical flow (Fig.~\ref{fig:pipeline}, right). Successive event slices within a small period of time are then inversely warped. Perfectly motion compensated slices should stack into a sharp profile, and we introduce a two-stage $Loss_{warp}$ to measure the warping quality. The sum of the losses $Loss=Loss_{warp}+w_{depth}Loss_{depth}+w_{mask}Loss_{mask}$ is backpropagated to train flow, inverse depth, and pose.

\subsection{Ego-motion Model}
 We assume that the camera motion is rigid with  a  translational velocity $v = (v_x,v_y,v_z)$ and a rotational velocity $\omega =(\omega_x, \omega_y, \omega_z)$, and   we also assume the camera to be calibrated. 
Let $\mathbf{X}=(X,Y,Z)^T$ be the world coordinates of a point and $\mathbf{x} = (x,y)^T$ be the corresponding image coordinates. The image velocity $\mathbf{u} = (u,v)^T$ is related to   $\mathbf{x}$,  the depth $Z$ and $t$ and $\omega$ as:
\begin{equation}
\resizebox{.9\hsize}{!}{$
\begin{pmatrix}
u\\v
\end{pmatrix}\\
=\frac{1}{Z}
\begin{pmatrix}
-1 &0 &x\\
0 &-1 &y
\end{pmatrix}
\begin{pmatrix}
v_x\\
v_y\\
v_z
\end{pmatrix}+
\begin{pmatrix}
xy &-1-x^2 &y\\
1+y^2 &-xy &-x
\end{pmatrix}
\begin{pmatrix}
\omega_x\\
\omega_y\\
\omega_z
\end{pmatrix}\\
= A \mathbf{p}
$}
\label{flow}
\end{equation}

Thus, for each pixel, given the inverse depth, there is a linear relation between the optical flow and the 3D  motion parameters 
%we can calculate the optical flow or pixel velocity using a simple matrix multiplication 
(Eq.~\ref{flow}). As it is common in the literature,  $\mathbf{p}$  denotes  the 3D motion (or pose vector) $(\mathbf{v},\omega)^T$, and $A$ here denotes  a $2\times6$ matrix. Due to scaling ambiguity in this formulation, depth $Z$ and  translation $(v_x, v_y, v_z)$ are computed up to a scaling factor. In our practical implementation, we normalize $Z$ by the average depth. 

We model the motion of individual moving objects as a 3D translation (without rotation), since most objects have relatively small size. The  motion (pose) of any object is modeled as the sum of the rigid background motion and the object translation.
Our network uses a mixture model for object segmentation - the 3D motion $p^i$ at a pixel $(x_i, y_i)$, is modeled as the sum of the camera motion $p_{ego}$ and  weighted  object translations, where the weights are obtained from motion masks as: 
\begin{equation}
p^i= p_{ego} + \sum_{j=1}^{C} m_j^i t_{j},
\label{pixel_pose}
\end{equation}
In the above equation $m_j^i$ are the motion mask weights for the $i-th$ pixel and $t_j$  the estimated translations of the $C$ objects.

\subsection {A Mixture Model for Ego-motion and Independently Moving Objects}

The pose network utilizes a mixture model to predict pixel-wise pose. 
%The network outputs a mixture of motion poses $(p_0,p_1,...,p_C)$ at the end of its encoding part. For the motion estimation problem, we make $p_0$ to be the ego-motion pose $p_{ego}$. We then let the network generate a few candidate residual poses $(r_1,...,r_C)$, and add them to $p_0$ to get the candidate relative poses with respect to the camera.
The network output of the encoder part is a set of poses $(p_0,t_1,...,t_C)$, where $p_0$ is the ego-motion pose $p_{ego}$, and $(t_1, ..., t_C)$ are   the translations \textit{with respect to the background} or residual translations. The residual translations are added to the ego-motion pose as in Eq.~\ref{pixel_pose} to get the candidate poses of objects relative to the camera. 
 %mixture of motion poses $(p_0,p_1,...,p_C)$ at the end of its encoding part. For the motion estimation problem, we make $p_0$ to be the ego-motion pose $p_{ego}$. We then let the network generate a few candidate residual poses $(r_1,...,r_C)$, and add them to $p_0$ to get the candidate relative poses with respect to the camera.
 
In the decoding part, the network predicts pixel-wise mixture weights or motion masks for the poses. We use the mixture weights and the  pose candidates to generate pixel-wise pose. The mixture weights sum to $1$ for each pixel.
We found experimentally that allowing a pixel to belong to multiple rigid motions as opposed to only one, leads to better results. This is because soft assignment allows the model to explain more directions of motions.
However, since  during training, the object masks are provided, qualitatively sharp object boundaries are learned.
 
 Using the mixture model representation allows us to differentiate object regions, moving with relatively small difference in 3D motion.
 
 %For a moving object that has a non-zero residual pose, the mixture model captures all the motion that can be linearly spanned by the candidate residual poses. It captures different directions by linearly combining the residual poses. It can also represent a same motion direction at a slower speed by uniformly reducing the weights assigned to the candidate residual poses and increasing the weight assigned to the ego-motion pose.   

\subsection{Loss functions}

We describe the loss functions used in the framework. It is noteworthy that the outputs of  our networks are multi-scale. The loss functions described in this section are also calculated at various scales. They are weighted by the number of pixels and summed up to calculate the total loss.

\subsubsection{Event Warping Loss}
\label{sec:wloss}

In the training process, we calculate
the optical flow and inversely warp events to compensate for the motion.
This is done by measuring the warping loss at two time scales, first for a rough estimate,  between slices, then for a refined estimate within a slice where we take full advantage of the timestamp information in the events.

Specifically, first using the optical flow estimate, we inversely warp neighboring slices to the center slice. To measure the alignment quality at the coarse scale, we take 3-5 consecutive  event slices, where each consists of $25$ milliseconds of motion information, and we use the absolute difference in event counts after warping as the loss: $$Loss_{coarse}=\sum_{-K\leq n\leq K,n\neq 0} |I^{warped}_n - I^{middle}|,$$
where $I^{warped}$ and $I^{middle}$ denote the three maps of positive events, negative events and average timestamps of the warped and the central slice, and $K$ is either 1 or 2.
To refine the alignment, we process the event point clouds and divide the slices into smaller slices of $1 ms$. Separately warping each of the small slices allows us to fully utilize  the time information contained in the continuous event stream.

We stack warped slices and use the following sharpness loss to estimate the warping quality. Intuitively speaking, if the pose is perfectly estimated, the stacking of inversely warped slices should lead to a motion-deblurred sharp image. Let $S=\sum_{n=-N}^N |I^{warped}_n|$ be the stacking of inversely warped event slices, where $n$ represents the $n$-th slice in a stack of $2N+1$ slices. Our basic observation is that the sparse quasi-norm $||\cdot||_p$ for  $0<p<1$ favors a sharp non-negative image over a blurred one. That is, $\sum_{i} |x_i|^p \geq (\sum_i |x_i|)^p$ for $0<p<1$. Based on this observation, we calculate the quasi-norm of $S$ to get the fine scale loss: $Loss_{fine}=||S||_{p},0<p<1$.

\subsubsection{Motion Mask Loss}
Given the ground truth motion mask, we apply a binary cross entropy loss on the mixture weight of the ego-motion pose component to constrain that our model applies the ego-motion pose in the background region: $Loss_{mask}=-\sum_{i \in background} log(m_0^i)$ To enforce that the mixture assignment is locally smooth, we also apply a smoothness loss on the first-order gradients of all the mixture weights.

\subsubsection{Depth Loss}
\label{sec:DepthLoss}
With ground truth depth available, we enforce the depth network output to be consistent with the ground truth. We adjust the network output and the ground truth to the same scale, which we denote as  $predict$ and $truth$ and apply the following penalty on their deviation: $Loss_{depth}=max(\frac{truth}{predict},\frac{predict}{truth})+\frac{|predict-truth|}{truth}$. Additionally, we apply a smoothness penalty on the second-order gradients of the prediction values, $Loss_{depth\_smooth}=||\Delta predict||_1$.

\subsection {Evenly Cascading Network Architecture}

%Explain the architecture high-level, we will also make a new figure.
We adopt the low parameter evenly cascaded convolutional network (ECN) architecture as our backbone network design~\cite{ye2018unsupervised}, and adapted it to an encoder decoder network borrowing concepts from the U-net design. The ECN network aggregates multilevel feature streams to make predictions. The low level features (Fig.~\ref{fig:pipeline}, light blue blocks) are scaled with bilinear interpolation and improved throughout the whole encoding-decoding structure via residual learning. Along that, the network also progressively generates high level features (Fig.~\ref{fig:pipeline}, darker blue blocks) in the encoding stage. The decoding stage proceeds reversely, the high level features are transformed by convolution and progressively merged back to the low level features to enhance them. Skip links (white arrows) are used in the network to connect features of the same shape, as in the original U-Net~\cite{DBLP:conf/miccai/RonnebergerFB15}.

\subsection {Prediction of Depth and Component Weights}

In the decoding stage, we make predictions using features at different resolutions and levels (Fig. ~\ref{fig:pipeline}). Initially, both high and low-level coarse features are used to predict a backbone prediction map. The prediction map is then upsampled and merged into existing feature maps  for refinements in the remaining decoding layers. In the middle stage, high level features as well as features in the encoding layers are merged into the low level features to serve as modulation streams. The  enhanced lower level features are used to estimate the prediction residue, which are usually also low-level structures. The residue is added to the current prediction map to refine it. The final prediction map is therefore obtained through successive upsamplings and refinements.

\section{EV-IMO Dataset}
\begin{figure}
\centering
\includegraphics[width=1.\columnwidth]{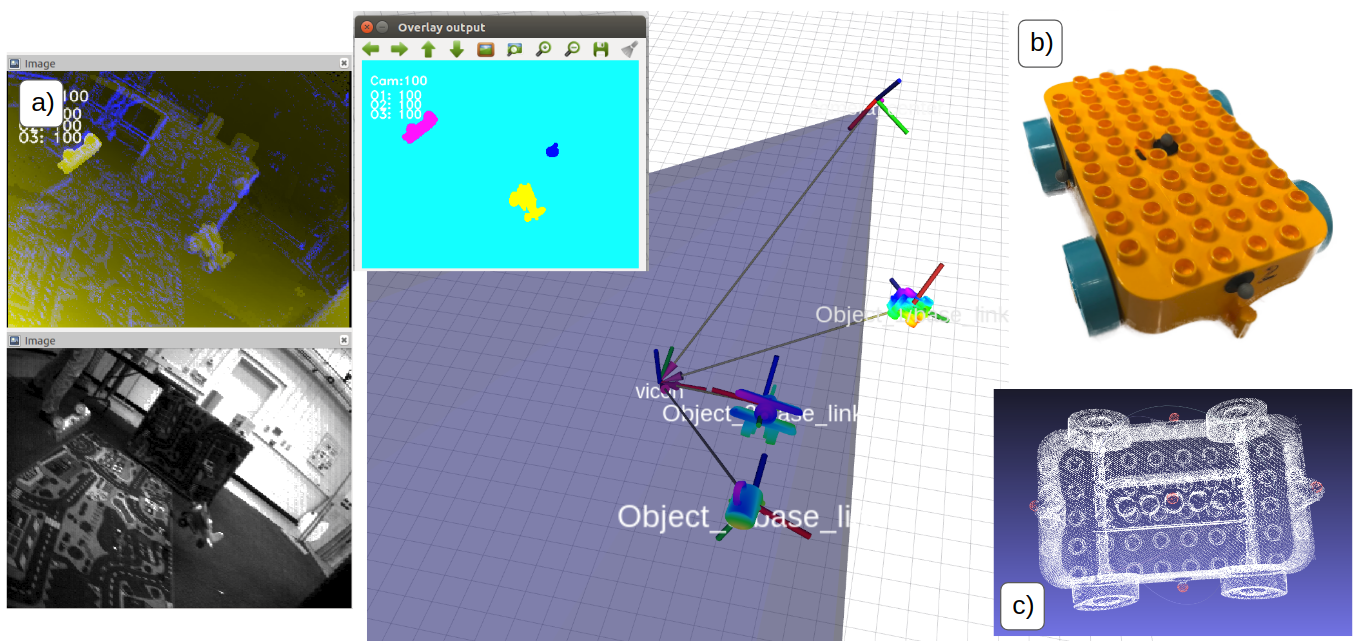}
\caption[The main interface of the automatic annotation tool]{\textit{\small{a) - The main interface of the automatic annotation tool. Camera cone of vision, depth and motion masks are visible. b) - Example object used in the dataset. c) 3D scan of the object.}}}
\vspace{-1.0\baselineskip}
\label{fig:interface_example}
\end{figure}

One of the contributions of this work is the collection of the \textit{EV-IMO} dataset - the first event camera dataset to include multiple independently moving objects and camera motion (at high speed), while providing accurate depth maps, per-object masks and trajectories at over 200 frames per second. The next sections describe our automated labeling pipeline, which allowed us to record more than 30 high quality sequences with a total length of half an hour. The source code for the dataset generation will be made available, to make it easier to expand the dataset in the future. A sample frame from the dataset is shown in Fig. \ref{fig:interface_example}.

%\begin{figure}
%\centering
%\includegraphics[width=0.9\columnwidth]{img/dataset_sample_0.png}
%\caption{\textit{\small{An example frame from the EV-IMO dataset. The event %information is overlaid on the depth ground truth on the left image, the %per-object masks are provided on the right image.}}}
%\vspace{-1.0\baselineskip}
%\label{fig:dataset_sample}
%\end{figure}

\subsubsection{Methodology}
Event cameras such as the DAVIS are designed to capture high speed motion and work in difficult lighting conditions. For such conditions classical methods of collecting depth ground truth, by calibrating a depth sensor with the camera, are extremely hard to apply - the motion blur from the fast motion would render such ground truth unreliable. Depth sensors have severe limitations in their frame rate as well. Furthermore it would be impossible  to automatically acquire object masks - manual (or semi-automatic) annotation would be necessary. To circumvent these issues we designed a new approach: 

\begin{enumerate}
  \item A static high resolution 3D scan of the objects, as well as 3D room reconstruction is performed before the dataset recording takes place.
  \item The VICON\textsuperscript{\tiny\textregistered} motion capture system is used to track both the objects and the camera during the recording.
  \item The camera center as well as the object and room scans are calibrated with respect to the  VICON\textsuperscript{\tiny\textregistered} coordinate frame.
  \item For every pose update from the VICON motion capture, the 3D point clouds are transformed and projected on the camera plane, generating the per-pixel mask and ground truth depth.
\end{enumerate}

This method allows to record accurate depth at  very high frame rate, avoiding the problems induced by frame-based collection techniques. While we acknowledge that this approach requires expensive equipment, we argue that our method is superior for event-based sensors, since it allows to acquire the ground truth at virtually any event time stamp (by interpolating poses provided at 200 Hz) - a property impossible to achieve with manual annotation.

\subsubsection{Dataset Generation}

Each of the candidate objects (up to 3 were recorded) were fitted with VICON\textsuperscript{\tiny\textregistered} motion capture reflective markers and 3D scanned using an industrial high quality 3D scanner. We use RANSAC to locate marker positions in the point cloud frame, and with the acquired point correspondences we transform the point cloud to the world frame at every update of the VICON. To scan the room, we place reflective markers on the Asus Xtion RGB-D sensor and use the tracking as an initialization for global ICP alignment.

To compute the position of the DAVIS camera center in the world frame we follow a simple calibration procedure, using a wand that is tracked by both VICON and camera. The calibration recordings will be provided with the dataset. 
The static pointcloud is then projected to the pixel coordinates $(x,y)$ in the camera center frame following equation \ref{eq:projecting_points}:

\begin{equation}
(x,y,1)^T = K  C  P_{davis}^{-1}  P_{cloud}  X_i
\label{eq:projecting_points}
\end{equation}
Here, $K$ is the camera matrix, $P_{davis}$ is a $4\times4$ transformation matrix between reflective markers on the DAVIS camera and the world, $C$ is the transformation between reflective markers on the DAVIS and the DAVIS camera center, $P_{cloud}$ is the transformation between markers in the 3D pointcloud and reflective markers in the world coordinate frame, and $X_i$ is the point in the 3D scan of the object.

Or dataset provides high resolution depth, pixel-wise object masks and accurate camera and object trajectories. We additionally compute, for every depth ground truth frame, the instantaneous camera velocity and the per-object velocity in the camera frame, which we use in our evaluations. We would like to mention, that our dataset allows to set varying ground truth frame rates - in all our experiments we generated ground truth at 40 frames per second.

\subsection{Sequences}
A short qualitative description of the sequences is given in Table \ref{table:seq_descr}. We recorded 6 sets, each consisting of 3 to 19 sequences. The sets differ in the background (in both depth and the amount of texture), the number of moving objects, motion speeds, and lighting conditions.

\begin{figure}
\centering
\includegraphics[width=1.\columnwidth]{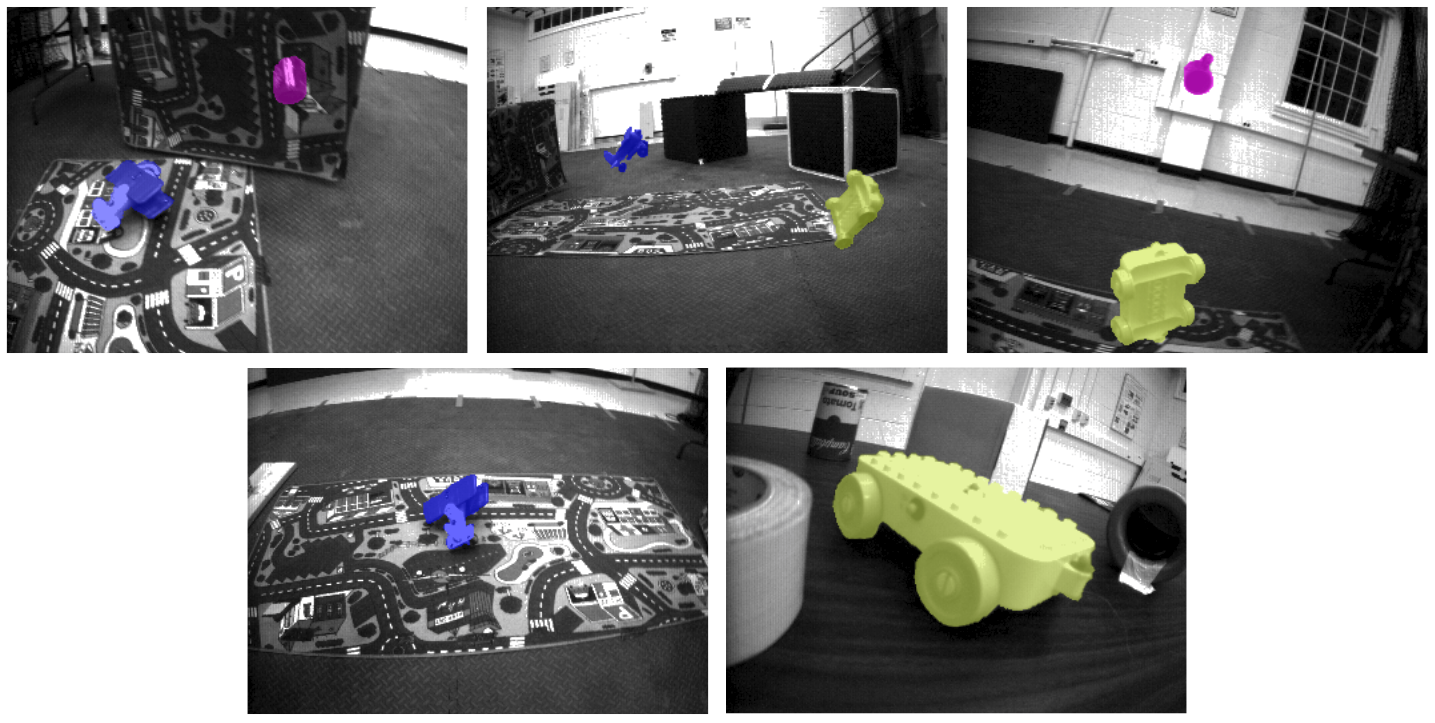}
\caption[Types of background geometry featured in the EV-IMO dataset ]{\textit{\small{Types of background geometry featured in the EV-IMO dataset(from left to right): 'table', 'boxes', 'plain wall', 'floor' and 'tabletop'.}}}
\vspace{-1.0\baselineskip}
\label{fig:backgrounds}
\end{figure}

\textbf{\textit{A note on the dataset diversity:}} It is important to note, that for  event-based cameras (which capture only edge information of the scene) the most important factor of diversity is the variability on \textit{motion}. Different motions create 3D event clouds which vary significantly in their structure, even with similar backgrounds. Nevertheless, we organize our sequences into four background groups - \textit{'table'}, \textit{'boxes'}, \textit{'plain wall'} and \textit{'floor'} (see Fig \ref{fig:backgrounds}), with the latter two having varying amounts of texture - an important factor for event cameras. We also include several tabletop scenes, with clutter and independently moving objects.

\renewcommand{\arraystretch}{1.3}
\begin{table}[h]
\caption[EV-IMO sequences]{\small{EV-IMO sequences}}
%\vspace{-1.0\baselineskip}
\begin{center}
\resizebox{1.0\columnwidth}{!}{\begin{tabular}{@{\extracolsep{5pt}}lcccccc}
\hline
& background & speed & texture & occlusions & objects & light \\
\hline
\textit{Set 1} & boxes      & low    & medium & low    & 1-2   & normal \\
\textit{Set 2} & floor/wall & low    & low    & low    & 1-3   & normal \\
\textit{Set 3} & table      & high   & high   & medium & 2-3 & normal \\
\textit{Set 4} & tabletop   & low    & high   & high   & 1      & normal \\
\textit{Set 5} & tabletop   & medium & high   & high   & 2     & normal \\
\textit{Set 6} & boxes      & high   & medium & low    & 1-3    & dark / flicker \\
\hline

\end{tabular}}
\end{center}
\vspace{-1.0\baselineskip}

\label{table:seq_descr}
\end{table}
\renewcommand{\arraystretch}{1}

\section{Experiments}
Learning motion segmentation on event-based data is challenging because the data from event-based sensors is extremely sparse (coming only from object edges). Nevertheless, we were able to estimate the full camera egomotion, a dense depth map, and the 3D linear velocities of the independently moving objects in the scene.

We trained our networks with the Adam optimizer using a starting learning rate of $0.01$ with cosine annealing for 50 epochs. The batch size was 32. We distributed the training over $4$-Nvidia GTX 1080Ti GPUs and the training finished within 24 hours. Inference runs at over 100 fps on a single GTX 1080Ti. 

In all  experiments, we trained on \textit{'box'} and \textit{'floor'} backgrounds, and tested on \textit{'table'} and \textit{'plain wall'} backgrounds (see Table \ref{table:seq_descr} and Fig. \ref{fig:backgrounds}). For the Intersection over Union (IoU) scores, presented in Table \ref{table:mo_est} the inferenced object mask was thresholded at $0.5$.

Our baseline architecture contains approximately 2 million parameters. It has  32 initial hidden channels and a growth rate of 32. The feature scaling factors are $\frac{1}{2}$ and $2$ for the encoding and decoding. Overall the networks have 4 encoding and 4 decoding layers. 

However, for many applications (such as autonomous robotics), precision is less important than computational efficiency and speed. We trained an additional shallow network with just 40 thousand parameters. In this setting we have  8 initial hidden channels and a growth rate of 8. The feature scaling factors are $\frac{1}{3}$ and $3$ respectively. The resulting networks have only 2 encoding and 2 decoding layers. We found that the 40k network is not capable of predicting object velocity reliably, but it produces reasonable camera egomotion, depth and motion masks, which can be tracked to extract the object translational velocities.

\subsubsection{Qualitative Evaluation}
\begin{figure}
\centering
\includegraphics[width=1.\columnwidth]{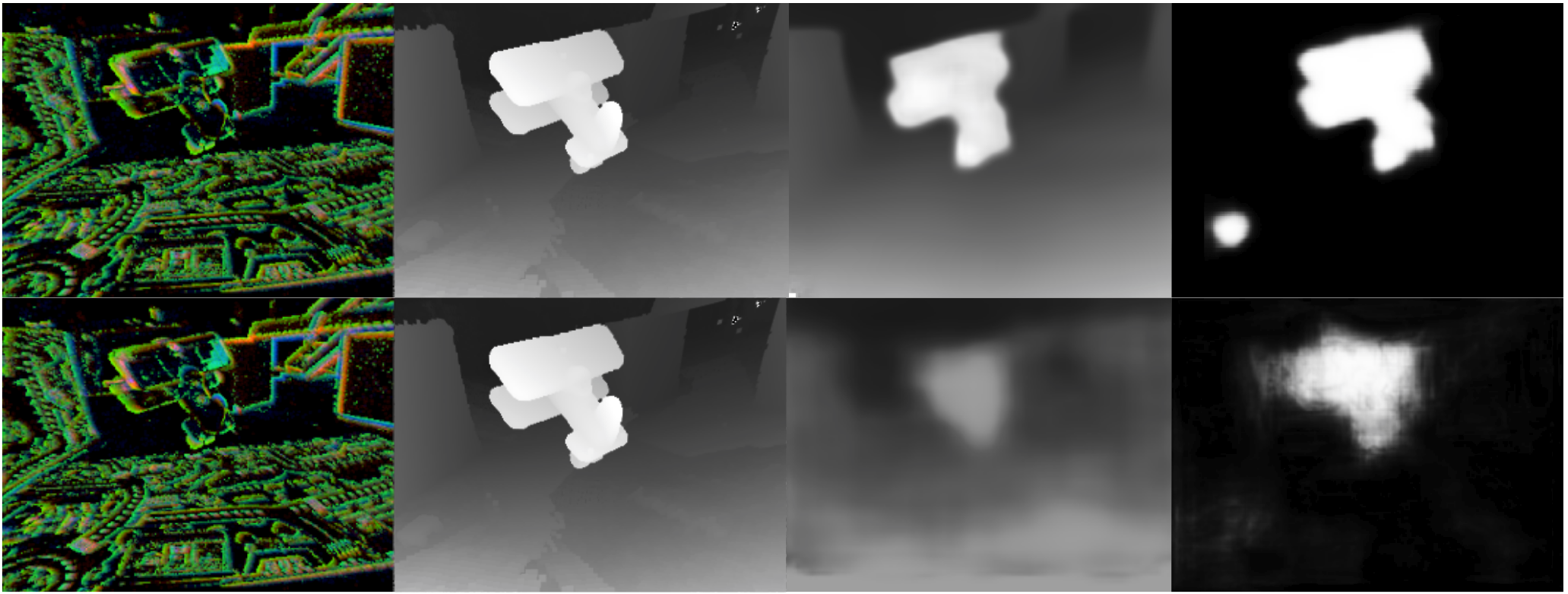}
\caption[Comparison of the full network inference quality with the small version]{\textit{\small{Comparison of the full network inference quality (2M parameters, top row) with the small version (40k parameters, bottom row)}}}
\vspace{-1.0\baselineskip}
\label{fig:big_vs_tiny}
\end{figure}

\begin{figure*}[t!]
\begin{center}
\includegraphics[width=0.95\textwidth]{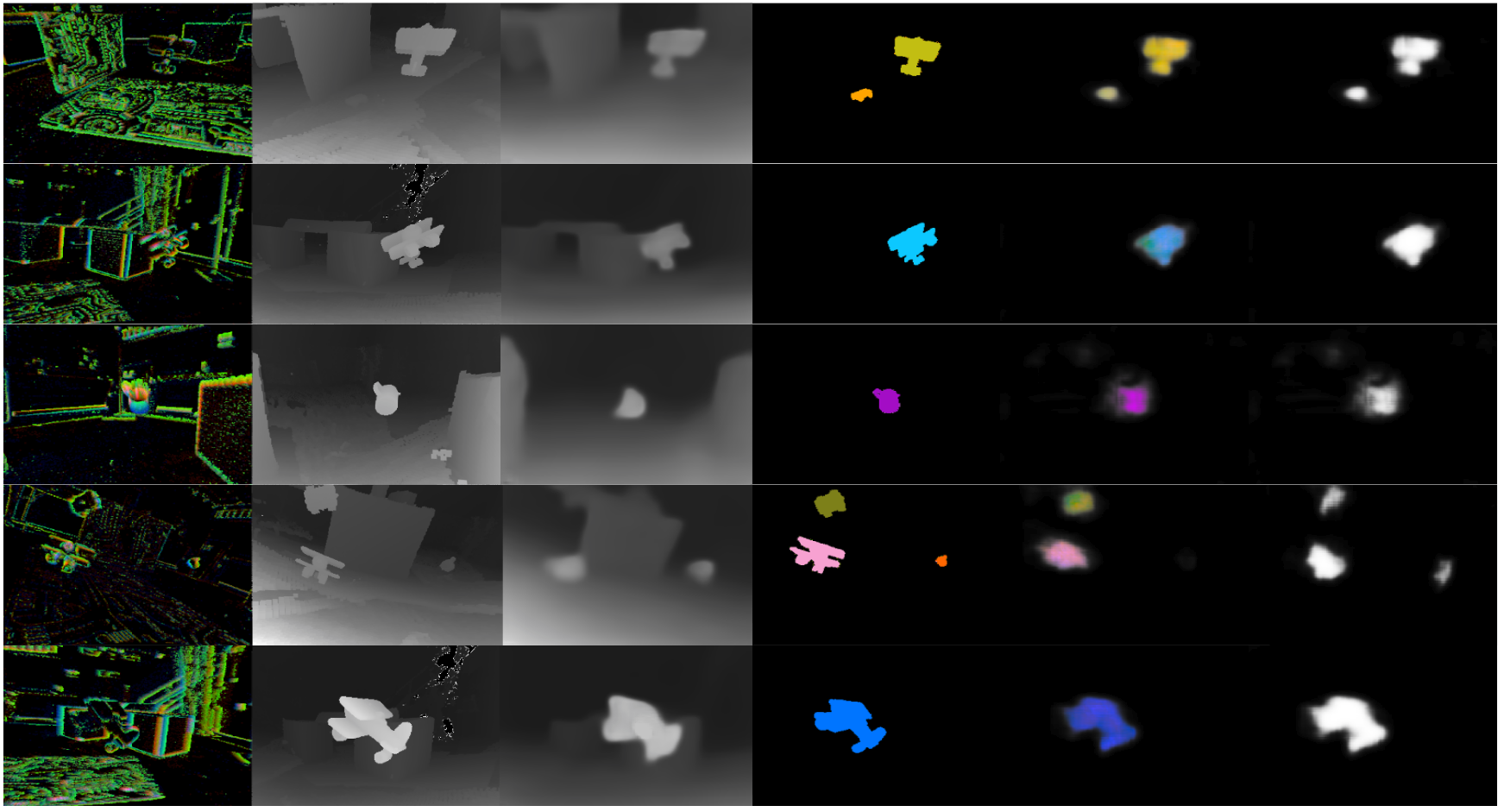}
\end{center}
\vspace{-0.5\baselineskip}
   \caption[Qualitative results from our evaluation]{\textit{\small{Qualitative results from our evaluation. The table entries from left to right: DVS input, ground truth for depth, network output for depth, ground truth pixel-wise pose, predicted pixel-wise pose, predicted motion mask. Examples were collected from \textit{EV-IMO} dataset. Best viewed in color.}}}
   \vspace{-1.0\baselineskip}
\label{fig:big_pic}
\end{figure*}

Apart from the quantitative comparison we present a qualitative evaluation in  Figs.
\ref{fig:big_pic} and \ref{fig:big_vs_tiny}. The per-object pose visualization (Fig. \ref{fig:big_pic}, columns 4 and 5) directly map  the 3D linear velocity to RGB color space.  The network is capable of predicting masks and pixel-wise pose in scenes with different amount of motion, number of objects or texture.

Fig. \ref{fig:big_vs_tiny} shows how the quality of the depth and motion mask output is affected by reducing the size of the network. While the background depth is affected only to a small degree, the quality of the object mask and depth suffers notably.

\subsubsection{Segmentation and Motion Estimation}

To evaluate the linear components of the velocities, for both egomotion and object motion, we compute the classical Average Endpoint Error (AEE). Since our pipeline is monocular, we apply the scale from the ground truth data in all our evaluations. To account for the rotational error of the camera (which does not need scaling) we compute the Average Relative Rotation Error $RRE = \|logm(R_{pred}^TR_{gt})\|_2$. Here $logm$ is the matrix logarithm, and $R$ are Euler rotation matrices. The $RRE$ essentially amounts to the total 3-dimensional angular rotation error of the camera. We also extract several sequences featuring fast camera motion and evaluate them separately. We present $AEE$ in m/s, and $RRE$ in radians/s in Table \ref{table:mo_est}.

We compute the averaged linear velocity of the independently moving objects within the object mask (since it is supplied by the network per pixel) and then also compute $AEE$. To evaluate the segmentation we compute the commonly used Intersection over Union (IoU) metric. Our results are presented in Table \ref{table:mo_est}.

\renewcommand{\arraystretch}{1.2}
\begin{table}[h]
\caption[Evaluation on segmentation and motion estimation]{\small{Evaluation on segmentation and motion estimation. The numbers in braces are values for the 40k version of the network. $AEE$ is in m/s, $RRE$ is in rad/s.}}
%\vspace{-1.0\baselineskip}
\begin{center}
\resizebox{1.0\columnwidth}{!}{\begin{tabular}{@{\extracolsep{5pt}}lcccc}
\hline
& Cam $AEE$  & Cam $RRE$ & Obj AEE & IOU \\
\hline
\textit{$table$} & 0.07 (0.09)  & 0.05 (0.08)   & 0.19   & 0.83 (0.63) \\
\textit{$plain wall$} & 0.17 (0.23)  & 0.16 (0.24)   & 0.38   & 0.75 (0.58) \\
\textit{$fast motion$} & 0.23 (0.28)  & 0.20 (0.26)   & 0.43   & 0.73 (0.59) \\
\hline

\end{tabular}}
\end{center}
\vspace{-1.0\baselineskip}

\label{table:mo_est}
\end{table}
\renewcommand{\arraystretch}{1}

\subsubsection{Comparison With Previous Work}
As there is no public code available for monocular SfM on event-based data, we evaluate on a 4-parameter motion-compensation pipeline~\cite{iROSBetterFlow}. We evaluated the egomotion component of the network on a set of sequences without IMOs and with no roll/pitch egomotion and with planar background found in \textit{'plain wall'} scenes, to make \cite{iROSBetterFlow} applicable (\cite{iROSBetterFlow} does not account for depth variation). Table ~\ref{table:classic} reports the results in m/s for the translation and in rad/s for the rotation. We were not able to achieve any meaningful egomotion results on scenes with high depth variation for \cite{iROSBetterFlow}.

\renewcommand{\arraystretch}{1.3}

\begin{table}[!htb]
\caption[Comparison of $EV-IMO$ with a classic method]{\small{Comparison of $EV-IMO$ with ~\cite{iROSBetterFlow}.}}
\vspace{0.5\baselineskip}

\centering
%\resizebox{.8\linewidth}{!}{
\begin{tabular}{lcc}
\hline
& $AEE$   & $RRE$\\
\hline\hline

\textit{EV-IMO} & 0.024  & 0.095    \\
\textit{Classical~\cite{iROSBetterFlow}} & 0.031  & 0.134    \\
\hline
\end{tabular}
%}
\label{table:classic}
\vspace{-1.5\baselineskip}

\end{table}
\renewcommand{\arraystretch}{1}

We also evaluate our approach against a recent method \cite{ye2018unsupervised} - \textit{ECN} network, which estimates optical flow and depth on the event-based camera output. The method was originally designed and evaluated on a road driving sequence (which features a notably more simple and static environment, as well as significantly rudimentary egomotion). Still, we were able to tune \cite{ye2018unsupervised} and train it on \textit{EV-IMO}. We provide the comparison for the depth for our baseline method, the smaller version of our network (with just 40k parameters) and \textit{ECN} in Table \ref{table:main_table_depth}.

\renewcommand{\arraystretch}{1.0}
\begin{table}[h]
\caption[Evaluation of the depth estimation]{\small{Evaluation of the depth estimation}}
\vspace{0.0\baselineskip}
\begin{center}
\resizebox{1.0\columnwidth}{!}{\begin{tabular}{@{\extracolsep{0pt}}lcccccc}
\hline

 & \multicolumn{3}{l}{Error metric} & \multicolumn{3}{l}{Accuracy metric} \\
 \cline{2-4}\cline{5-7}
 & Abs Rel & RMSE log & SILog & $\delta < 1.25$  & $\delta < 1.25^2$ & $\delta < 1.25^3$  \\

\hline\hline
\multicolumn{7}{c}{Baseline Approach} \\
\hline\hline
\textit{plain wall}      & 0.16 & 0.26 & 0.07 & 0.87 & 0.95 & 0.97 \\
\textit{cube background}  & 0.13 & 0.20 & 0.04 & 0.87 & 0.97 & 0.99 \\ 
\textit{table background}  & 0.31 & 0.32 & 0.12 & 0.74 & 0.90 & 0.95 \\ 
\hline\hline
\multicolumn{7}{c}{40k Network} \\
\hline\hline
\textit{plain wall}      & 0.24 & 0.33 & 0.11 & 0.75 & 0.90 & 0.95 \\
\textit{cube background}  & 0.20 & 0.26 & 0.07 & 0.77 & 0.92 & 0.97 \\ 
\textit{table background}  & 0.33 & 0.34 & 0.15 & 0.65 & 0.87 & 0.95 \\ 
\hline\hline
\multicolumn{7}{c}{ECN} \\
\hline\hline
\textit{plain wall}      & 0.67 & 0.59 & 0.33 & 0.27 & 0.52 & 0.80 \\
\textit{cube background}  & 0.60 & 0.56 & 0.30 & 0.29 & 0.53 & 0.78 \\ 
\textit{table background}  & 0.47 & 0.48 & 0.23 & 0.45 & 0.69 & 0.86 \\ 

\hline
\hline
\end{tabular}}
\end{center}
\vspace{-1.0\baselineskip}

\label{table:main_table_depth}
\end{table}
\renewcommand{\arraystretch}{1}

We conducted the experiments on sequences featuring a variety of backgrounds and textures (the lack of texture is a limiting factor for event-based sensors). Even though \textit{ECN} \cite{ye2018unsupervised} was not designed to segment independently moving objects, the comparison is valid, since it infers depth from a single frame. Instead, we attribute the relatively low performance of \cite{ye2018unsupervised} to a significantly more complex motion present in EV-IMO dataset, as well as more diverse depth background. 

\section{Conclusions}
Event-based sensing promises advantages over classic video processing in applications of motion estimation because of the data's unique properties of sparseness, high temporal resolution, and low latency. In this paper, we presented a compositional NN pipeline, which uses a combination of unsupervised and supervised components and is capable of generalizing well across different scenes. We also presented the first ever method of event-based motion segmentation with evaluation of both camera and object motion, which was achieved through the creation of a new state of the art indoor dataset - \textit{EV-IMO}, recorded with the use of a VICON\textsuperscript{\tiny\textregistered} motion capture system.

Future work will delve into a number of issues regarding the design of the NN and usage of event data. Specifically, we consider it crucial to study event stream augmentation using partially or fully simulated data. We also plan to investigate ways to include tracking and connect the estimation over successive time slices, and investigate different alternatives of including the grouping of objects into the pipeline.

\section{Acknowledgements}
The support of the Northrop Grumman Mission Systems University Research Program, ONR under grant award
N00014-17-1-2622, and the National Science
Foundation under grant No. 1824198 are
greatly acknowledged.
%--------------------------n --------------------------

%----------------------------------------------------
{\small
\bibliographystyle{ieee}
\bibliography{references}
}

\end{document}